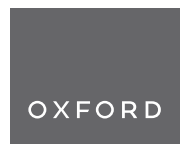

Data and text mining

# BERN2: an advanced neural biomedical named entity recognition and normalization tool

Mujeen Sung ⓘ [1,†], Minbyul Jeong ⓘ [1,†], Yonghwa Choi ⓘ [1], Donghyeon Kim ⓘ [2], Jinhyuk Lee ⓘ [1,*,‡] and Jaewoo Kang ⓘ [1,3,*,‡]

[1]Department of Computer Science and Engineering, Korea University, Seoul 02841, Republic of Korea, [2]AIRS Company, Hyundai Motor Group, Seoul 06620, Republic of Korea and [3]AIGEN Sciences, Seoul 04778, Republic of Korea

*To whom correspondence should be addressed.
†The authors wish it to be known that, in their opinion, the first two authors should be regarded as Joint First Authors.
‡The authors wish it to be known that, in their opinion, the last two authors should be regarded as Joint Last Authors.
Associate Editor: Karsten Borgwardt

## Summary

In biomedical natural language processing, named entity recognition (NER) and named entity normalization (NEN) are key tasks that enable the automatic extraction of biomedical entities (e.g. diseases and drugs) from the ever-growing biomedical literature. In this article, we present BERN2 (Advanced Biomedical Entity Recognition and Normalization), a tool that improves the previous neural network-based NER tool by employing a multi-task NER model and neural network-based NEN models to achieve much faster and more accurate inference. We hope that our tool can help annotate large-scale biomedical texts for various tasks such as biomedical knowledge graph construction.

**Availability and implementation:** Web service of BERN2 is publicly available at http://bern2.korea.ac.kr. We also provide local installation of BERN2 at https://github.com/dmis-lab/BERN2.

**Contact:** jinhyuk_lee@korea.ac.kr or kangj@korea.ac.kr

**Supplementary information:** Supplementary data are available at *Bioinformatics* online.

## 1 Introduction

Biomedical text mining is becoming increasingly important due to the constantly growing volume of biomedical texts. From these texts, biomedical entities of various types such as gene/protein or disease can be automatically annotated with named entity recognition (NER) and linked to concept unique IDs (CUIs) with named entity normalization (NEN). Many biomedical text mining tools combine NER and NEN in a single pipeline to support large-scale annotation of biomedical texts. One popular example is PubTator Central (PTC) (Wei *et al.*, 2019). With recent progress in biomedical language models (Gu *et al.*, 2022; Lee *et al.*, 2020), biomedical text mining tools that are more accurate than PTC have been introduced (Kim *et al.*, 2019) and they are often used for downstream tasks such as biomedical knowledge graph construction (Xu *et al.*, 2020) and biomedical search engine (Köksal *et al.*, 2020).

However, existing biomedical text mining tools for biomedical NER, which often come with NEN, have several limitations. First, they provide annotations for a small number of biomedical entity types (e.g. five entity types in Weber *et al.*, 2021). Second, they often use multiple single-type NER models to annotate entities of different types (Kim *et al.*, 2019), which require larger Graphics Processing Unit (GPU) memory for parallelization, but are very slow at inference when used sequentially. Lastly, many tools employ NEN models based on pre-defined rules with dictionaries (Kim *et al.*, 2019; Wei *et al.*, 2019), but they cannot cover complex morphological variations of biomedical entity mentions. For instance, a simple dictionary-matching NEN model cannot normalize 'oxichlorochine' into its canonical mention 'hydroxychloroquine' (mesh: D006886) unless the dictionary explicitly contains the mention 'oxichlorochine'.

As shown in Table 1, our proposed tool, BERN2 (Advanced Biomedical Entity Recognition and Normalization) addresses these challenges by (i) supporting nine biomedical entity types, which are the largest among other commonly used tools listed in the table, (ii) dramatically reducing the annotation time by using a single multi-task NER model and (iii) combining rule-based and neural network-based NEN models to improve the quality of entity normalization. We provide BERN2 as a web service with RESTful Application Programming Interface (API) and also allow users to locally install it. The usage of BERN2 is detailed in Supplementary Data A.

## 2 Materials and methods

BERN2 is designed to recognize and normalize nine types of biomedical entities (gene/protein, disease, drug/chemical, species,

**Table 1.** Comparison of different biomedical text mining tools

| Tool | NER →NEN | Supported types | Sec/abstract | |
|---|---|---|---|---|
| | | | Plain text | PMID |
| PTC | ML → Rule | Ge/Di/Dr/Sp/Mu/CL | N/A | 0.86 ± 0.20[a] |
| HUNFLAIR | Neural → N/A | Ge/Di/Dr/Sp/CL | 0.53 ± 0.24 | N/A |
| BERN | Neural → Rule | Ge/Di/Dr/Sp/Mu | 1.08 ± 0.31 | 1.43 ± 0.19 |
| BERN2 | Neural → Rule and Neural | Ge/Di/Dr/Sp/Mu/CL/CT/DNA/RNA | 0.33 ± 0.29 | 0.03 ± 0.09 |

*Note*: ML, machine learning models such as HMM, CRF, etc.; Rule, dictionary and rule-based models; Neural, neural network-based models; Ge, gene/protein; Di, disease; Dr, drug/chemical; Sp, species; Mu, mutation; CL, cell line; CT, cell type; Sec/Abstract, average time in second (± std.) taken to annotate a PubMed abstract; computed over randomly sampled 30K abstracts using local installation.

[a]Measured from web service due to the lack of local installation support.

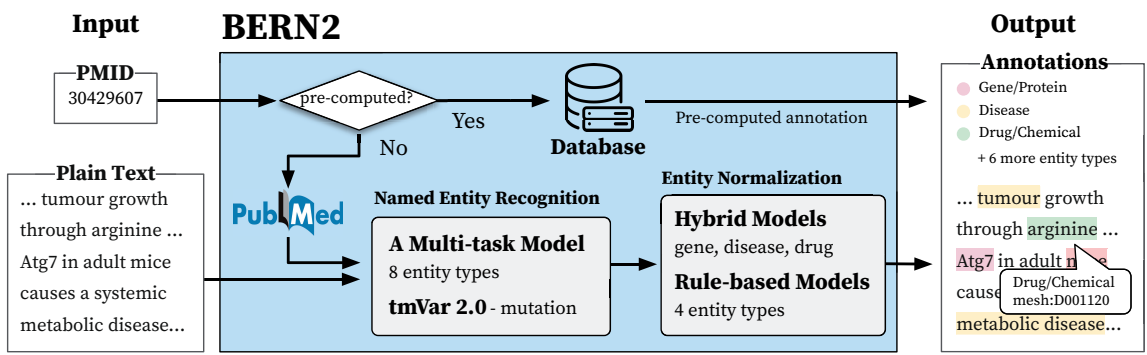

**Fig. 1.** An overview of BERN2. Given plain text or a PubMed ID (PMID), BERN2 recognizes nine biomedical entity types and normalizes each concept

mutation, cell line, cell type, DNA and RNA). As illustrated in Figure 1, we support two input formats: plain text and PubMed ID (PMID).

When plain text is given, a multi-task NER model of BERN2 first extracts the exact positions and types of biomedical named entities in the text (see Section 2.1 for a detailed description of the multi-task NER model). For example, for plain text '... tumor growth through arginine...,' our NER model locates the positions of 'tumor' and 'arginine' in the text and classifies them as disease and drug/chemical types, respectively. These named entities are then normalized into corresponding CUIs in designated dictionaries by our NEN models. We use hybrid NEN models for three entity types (gene/protein, disease and drug/chemical) to increase the number of entities being normalized correctly (see Section 2.2 for a detailed description of the hybrid NEN model). For example, our NEN model normalizes a gene/protein mention 'atg7' into 'NCBIGene: 10533' and a drug/chemical mention 'arginine' into 'mesh: D001120'.

We use the plain text annotation mode described above to pre-compute the annotations of PubMed articles (abstracts only), which are stored in an external database. Hence, when PMIDs are given, BERN2 returns pre-computed annotations from its database whenever possible, which is much faster than annotating the plain text as shown in Table 1. If an abstract of a PubMed article has not been pre-computed (i.e. not found in the database), BERN2 annotates it and stores the annotation results in the database.

### 2.1 Multi-task named entity recognition

While BERN (Kim *et al.*, 2019) employs accurate NER models based on a pre-trained biomedical language model (Lee *et al.*, 2020), it uses multiple single-type NER models (i.e. four BioBERT models to annotate four entity types except for mutation), which requires a large amount of GPU memory for parallelization but makes the entire pipeline slow without such parallel inference. BERN2 adopts a multi-task NER model that supports efficient parallel inference for eight entity types (except for mutation).

Following Wang *et al.* (2019), our multi-task NER model consists of a shared backbone model and a separate task-specific layer for each entity type. We use Bio-LM (Lewis *et al.*, 2020), a state-of-the-art pre-trained biomedical language model, as our backbone model and use two-layer MLP with ReLU activation as a task-specific layer. Each task-specific layer outputs probabilities of whether each token is the beginning, inside or outside (BIO) of named entities. During training, we merge five training sets of all entity types. We use BC2GM (Smith *et al.*, 2008) for gene/protein, NCBI-disease (Doğan *et al.*, 2014) for disease, BC4CHEMD (Krallinger *et al.*, 2015) for drug/chemical, Linnaeus (Gerner *et al.*, 2010) for species and JNLPBA (Kim *et al.*, 2004) for cell line, cell type, DNA and RNA.

At inference, our NER model takes an input text and outputs predictions from all task-specific layers in parallel. In fact, the multi-task NER model in BERN2 only consumes a small amount of GPU memory that is on par with using a single pre-trained LM since having multiple task-specific layers—two-layer MLP each—only adds a small number of parameters. This also allows us to adopt a larger pre-trained language model on a single GPU, which often outperforms smaller models. In this work, we use the large version of Bio-LM. (We use the *RoBERTa-large-PM-M3-Voc* checkpoint from https://github.com/facebookresearch/bio-lm.) Compared to BERN whose NER model has approximately 432M parameters (i.e. four LMs with 108M parameters each) and takes 600 ms to annotate four entity types, our multi-task NER model in BERN2 has approximately 365M parameters (i.e. a single LM with 365M parameters) and takes 200 ms to annotate eight entity types while also enjoying the expressiveness of a large pre-trained LM. Note that due to the lack of public training sets for mutations, both BERN and BERN2 use tmVar2.0 (Wei *et al.*, 2018) for mutation NER. We refer interested readers to Supplementary Data B for a detailed description of the NER model used by BERN2.

### 2.2 Hybrid named entity normalization

Rule-based NEN models that are often used by biomedical text mining tools (Kim *et al.*, 2019) cannot handle all morphological

**Table 2.** Results on biomedical NER benchmarks

| Dataset | Type | PTC | HUNFLAIR | BERN | BERN2 |
|---|---|---|---|---|---|
| BC2GM | Gene/protein | 78.8 | 77.9 | 83.4 | **83.7** |
| NCBI-disease | Disease | 81.5 | 85.4 | 88.3 | **88.6** |
| BC4CHEMD | Drug/chemical | 86.7 | 88.9 | 91.2 | **92.8** |
| tmVar2 | Mutation | **93.7** | N/A | **93.7** | **93.7** |
| Linnaeus | Species | 85.6 | **93.2** | 88.0 | 92.7 |
| JNLPBA | Cell line | N/A | 64.9 | N/A | **78.6** |
|  | Cell type | N/A | N/A | N/A | **80.7** |
|  | DNA | N/A | N/A | N/A | **77.8** |
|  | RNA | N/A | N/A | N/A | **76.5** |

*Note*: F1 score is reported.

**Table 3.** Results on biomedical NEN benchmarks

| Dataset | Type | PTC[a] | BERN | BioSyn[b] | BERN2 |
|---|---|---|---|---|---|
| BC2GN | Gene/protein | 93.8 | 93.8 | 91.3 | **95.9** |
| BC5CDR | Disease | 88.9 | 90.7 | 93.5 | **93.9** |
|  | Drug/chemical | 94.1 | 92.8 | **96.6** | **96.6** |

*Note*: Accuracy is reported.
The highest scores in each row are boldfaced.
[a]Wei *et al.* (2015) for BC2GN and Leaman and Lu (2016) for BC5CDR.
[b]A single NEN model (not a text mining tool) evaluated on each test set.

variations of biomedical named entities. Instead, Sung *et al.* (2020) introduce BioSyn, a neural network-based biomedical NEN model that leverages vector representations of entities to cover such variations. Specifically, BioSyn builds a dictionary embedding matrix from an entity encoder, where each row vector denotes the representation of an entity name in the dictionary. The input mention embedding is computed from the same entity encoder and BioSyn retrieves an entity name from the dictionary matrix that has the highest inner product score with the input mention embedding. Each mention is then normalized into the CUI of the retrieved entity.

BERN2 first tries rule-based normalization on each named entity and only the ones that were not normalized by the rule-based models are then normalized by BioSyn. We employ this hybrid approach for three entity types (gene/protein, disease and drug/chemical) where fine-tuned BioSyn is available. (We use the checkpoints *biosyn-sapbert-bc2gn* for gene/protein, *biosyn-sapbert-bc5cdr-disease* for disease and *biosyn-sapbert-bc5cdr-chemical* for drug/chemical from https://github.com/dmis-lab/BioSyn.) Other types are processed by rule-based models. With our hybrid NEN model, we safely increase the number of correctly normalized entities. For the sake of users, we also show whether each entity has been normalized by the rule-based NEN model or BioSyn in the annotation results. Supplementary Data C provides details of our normalization model.

## 3 Results

### 3.1 Named entity recognition

Table 2 shows the NER performance of different biomedical text mining tools including PTC (Wei *et al.*, 2019), HUNFLAIR (Weber *et al.*, 2021), BERN (Kim *et al.*, 2019) and BERN2. While BERN2 supports the largest number of entity types, it also outperforms other tools on most types except on the species type.

### 3.2 Named entity normalization

Table 3 shows the normalization accuracy of the BC2GN (gene/protein) and BC5CDR (disease and drug/chemical) test sets. Again, BERN2 that uses hybrid NEN (rule-based + BioSyn) outperforms other tools.

## 4 Conclusion

In this article, we present BERN2, a biomedical text mining tool for accurate and efficient biomedical NER and NEN. With a multi-task NER model and hybrid NEN models, BERN2 outperforms existing biomedical text mining tools while providing annotations more efficiently. We support both web service and local installation of BERN2 for the ease of employing BERN2 in other systems. Supplementary Data D provides example usages where BERN can be easily replaced with BERN2.

## Funding

This work was supported in part by National Research Foundation of Korea (NRF-2020R1A2C3010638, NRF-2014M3C9A3063541), the Ministry of Health & Welfare, Republic of Korea (HR20C0021) and the ICT Creative Consilience program (IITP-2021-0-01819) supervised by the IITP (Institute for Information & communications Technology Planning & Evaluation).

*Conflict of Interest*: none declared.